# Counter Hate Speech in Social Media: A Survey

Dana Alsagheer, Hadi Mansourifar, Weidong Shi

{dralsagh,hmansour,wshi3}@central.uh.edu

*Abstract*— With the high prevalence of offensive language against minorities in social media, counter-hate speeches (CHS) generation is considered an automatic way of tackling this challenge. The CHS is supposed to appear as a third voice to educate people and keep the social [red lines bold] without limiting the principles of freedom of speech. In this paper, we review the most important research in the past and present with a main focus on methodologies, collected datasets and statistical analysis CHS's impact on social media. The CHS generation is based on the optimistic assumption that any attempt to intervene the hate speech in social media can play a positive role in this context. Beyond that, previous works ignored the investigation of the sequence of comments before and after the CHS. However, the positive impact is not guaranteed, as shown in some previous works. To the best of our knowledge, no attempt has been made to survey the related work to compare the past research in terms of CHS's impact on social media. We take the first step in this direction by providing a comprehensive review on related works and categorizing them based on different factors including impact, methodology, data source, etc.

*Index Terms*— Counter Speech, Hate Speech Detection, Reactance

## I. Introduction

ROM misogyny and homophobia to xenophobia and racism, online hate speech has become a topic of greater Fconcern as the Internet matures, particularly as its offline impacts become more widely known. With hate-fueled tragedies across the US and New Zealand, 2019 has seen a continued rise in awareness of how social media and fringe websites are being used to spread hateful ideologies and to instigate violence [1][2]. As a result, the CHS is supposed to appear as a third voice to educate people and keep the social red lines bold without limiting the principals of freedom of speech. In this paper, we study the methods that were used in papers to detect hate speech and study the impact of CHS. For detecting the CHS two methods were used:

Hard Counter Hate Speech and (ii) Soft Counter Hate Speech. Both of them are defined in detail in the next subsections.

### A. Hard Counter Hate Speech

The online spreading of hate speech caused several countries and companies to perform laws against hate speech to prevent citizens from this behavior. There are laws that ban hate speech implemented in some countries such as Germany, the United States of America, and France. The responses of the hate speech for different platforms such as Twitter and Facebook usually will be the deletion of the comments, messages or the user account. Although these laws may lead to reducing the impact of hate speech in online media, they will harm freedom of speech. Another effective prospective way to change the mentality of hate speakers is called CHS which mainly adds more speech to their conversation [3]. The Council of Europe supports an initiative called 'No Hate Speech Movement' to diminish the levels of acceptance of hate speech and also to develop online youth participation and citizenship, including in Internet governance processes [3]. Countering Online Hate Speech is a study done by UNESCO the main purpose of this study is to guide the country on how to deal with this problem with no harm to freedom of speech [3]. Social media sites, such as Facebook have started CHS programs to tackle hate speech. Facebook has clearly announced that CHS has the potential to have positive effects and is more likely to succeed in the long run [4]. The CHS has massive benefits such as it is fast, flexible, and reacting. Moreover, it is qualified to deal with the extremism in different languages and causes no harm to freedom of speech.

### B. Soft Counter Hate Speech

According to [5], soft CHS is defined as all kinds of public speeches which can diminish inter-group violence while protecting the rights of freedom of expression. In the natural language processing context, the CHS generation is a recent investigated recently [6]. In this system, automatically generated CHS is supposed to be posted as a set of new comments in order to provide an extra positive voice to keep the overall language less dominated by hateful comments. The main assumption in CHS generation is that any attempt to insert new positive comments automatically can be helpful without measuring the generated CHS impact.

*C. Generation of Counter Hate Speech*

The main idea is to directly interfere in the discussion with textual responses to counter the hate content and to prevent it from further spreading by using automation strategies, such as natural language generation. The core of the study is to examine methods to obtain high-quality a Counter-Narrative (CN) which is a non-aggressive response that offers feedback through fact-bound arguments and reduces the efforts from experts. Starting by comparing data collection methods to ensure that datasets must meet two major requirements: (i) data quantity (ii) data quality. The generation of an effective automatic CN depends on the right trade-off between the two mentioned requirements [7].

The rest of the paper is organized as follows. Section II demonstrates the research method. Section III studies the related works. Section IV presents the methodology. Section V demonstrates data collection. Section VII provides the evaluation experimental. Section study VII challenges and opportunities. CHS in the media and the negative impact. Section IX. review the paper and finally section X concludes the paper.

## II. RESEARCH METHODS

This study focuses on the questions of what the impact of CHS on social media is. We decided in favor of using this term since it can be considered a wide range term. Our method for choosing the research to study was to follow the Systematic Literature Review steps, which are to identify and interpret relevant research to answer the research question [8]. Our approach methodology included the following steps:

1) Identify the purpose and intended goals of the review
2) Searching for literature to explain and justify the comprehensiveness of the search
3) A comprehensive study of screening for inclusion and exclusion for each reference
4) We extract data from each study to include the following variables: the impacts of CHS, data collections method, generation, and classification.
5) Analyze and summarize the of result of each study
6) Writing the review about the findings

## III. RELATED WORKS

Hate speech detection as a general problem has been investigated before in the research community. CHS is currently advocated by social networks as a measure for delimiting the effects of hate speech.

Mathew et al. [9] proposed a model which predicts if a given Twitter user is a hateful or CHS account with an accuracy of 78%. The model was provided a data set containing 1290 tweet-reply pairs in which the tweets are hate-speech, and their replies are CHS.

To study the impact of the CHS Strachan et al. [10] studied the evidence for positive impact in the programming interventions on countering hate speech; the change of behavior should be studied in long-term to be effective and they blamed this for the lack of evidence. Finally, the author provides different examples of proven successful interventions.

Bartlett et al. [4] attempted to study the effect of CHS on Facebook. The author believes in the importance that the principle of internet freedom should be maintained and that it should be a place where people feel they can speak their mind openly and freely. Although the author is against the removal of content, including when dealing with extreme or radical content, he believed that this can and should be put on an empirical basis to help us better understand the phenomena and how to respond. Also, they conducted a study in separate regions; in the order they were carried out: France, UK, Morocco and Tunisia, Indonesia, and India. Depending on the region, the context in which CHS is produced and shared changes and CHS content would operate differently.

Sponholz et al. [11] study the role of CHS in media controversies surrounding hate speech. The main contribution is analyzing the case of the Italian best-selling author Oriana Fallaci. She published the book *The Rage and the Pride* after the terror attack against the Twin Towers which was full of hate speech toward Muslims such as "Muslims breed like rats". They conclude that CHS did not lead either to consensus or to the refutation of such but increased the tone of the debate and added to media operational bias.

Schieb et al. [12] have proven that CHS can have a considerable impact on a certain audience. To study defining factors for the success of CHS, they set up a computational simulation model that is used to answer general questions concerning the effects that hinder or support the impact of CHS due to the lack of availability of data.

Ernst et al. [13] analyzed the random samples on user comments beneath videos tagged with whatIS, which have been released on YouTube within the framework of the CHS campaign Begriffswelten. They studied if the counter messages can create critical awareness for the topics presented in such videos as a positive outcome on stereotypical representations and discussions generating biased opinions towards Islam respectively and Muslims in Germany. The research also considers the negative potential of counter messages to pave the way for hate speech.

Chung et al. [14] provide a large-scale, multilingual, expert-based data set of hate speech counter narrative pairs. This dataset has been built with the effort of 100 operators of three different NGOs that applied their training and expertise to the task.

Mathew et al. [15] analyzed the results in various interesting insights such as the CHS comments receive double the likes received by the non-CHS comments. For certain communities' the majority of the non-CHS comments tend to be hate speech, the different types of CHS are not all equally effective and the language choice of users posting CHS is largely different from those posting non-CHS as revealed by a detailed psycholinguistic analysis. What is interesting in their study is using machine learning models to detect CHS in YouTube videos with an F1-score of 0.73. Also, they provide a data set for CHS by using comments from YouTube to perform a measurement study characterizing the linguistic structure of CHS.

Tekiroglu et al. [7] studied how to collect responses to hate effectively, utilizing large-scale unsupervised language models, and the best annotation strategies/neural architectures that can be used for data filtering before expert validation/post-editing. They mentioned a strategy that data sets must meet two criteria: quality and quantity. They studied how to collect responses to hate effectively, utilizing large-scale unsupervised language models, and the best annotation strategies/neural architectures that can be used for data filtering before expert validation/post-editing.

The case study done by Garland et al. [16] details how people in different countries react to CHS content on Facebook and works to identify what types of content were most likely to be engaged with by users. The study shows that users engage with CHS depending on their location in the world which indicates there is no broad approach that covers the entirety of Facebook. It requires specific approaches for locations and countries for which Facebook can provide an important platform to spread messages confronting IS narratives and ideology. Also, [17] they analyzed the effectiveness of CHS using several different macro-and micro-level measures to analyze 180,000 political conversations that took place on German Twitter over four years. findings suggest that organized hate speech is associated with changes in public discourse and that CHS especially when organized—may have a positive impact on the hateful rhetoric in online discourse. Table I illustrates the categories of CHS that used in each research. Finally, we study how frequently each recherche is cited by other research Figure I illustrates the relation between each research. For instance, [15] has been cited by both [16] and [17].

## IV. METHODOLOGIES

In this section, we review the methodologies used for CHS analysis, classification and generation.

*A. Features*

Before investigating the technical approaches, we need to review the features selected by researchers for CHS analysis, classification and generation [15].

- TF-IDF: Term Frequency-Inverse Document Frequency vectors.
- User profile properties: We use several user account properties such as # favorites per day, # tweets per day, # followers per day, # friends per day, # listed per day, and whether the account is verified.
- BoW: approach uses the average of the Global Vectors for Word Representation (GloVe) word embeddings to represent a sentence [15].
- Lexical properties: Lexical semantics also known as lexicosemantics, as a subfield of linguistic semantics, is the study of word meanings. It includes the study of how words structure their meaning, how they act in grammar and compositionality and the relationships between the distinct senses and uses of a word [15].
- Affect properties: For each user account we calculate the average sentiment, profanity, and subjectivity expressed in the tweet history [17].

*B Classification methods*

In this section, we review the classification approaches used in previous CHS research.

1) Support Vector Machine: a supervised learning model with associated learning algorithms that analyze data for pattern classification. It can be applied to many classifications' patterns such as image recognition, speech recognition, text categorization, face detection, and faulty card detections. SVM models are algorithmic implementations of ideas from statistical learning theory [18][19]. The main idea of pattern classification is based on prior knowledge or statistical information extracted from raw data. SVM builds optimal separating boundaries between data sets by solving a constrained quadratic optimization problem b by using different kernel functions, modifying degrees of non-linearity and flexibility which can be included in the model [19][20]. The advantage of statistical learning theory is the outline for studying the problem of gaining knowledge, making predictions, and decisions from a data set [21][22]. Although SVM builds optimal separating boundaries between data sets by solving a constrained quadratic optimization problem, it does not provide a probabilistic explanation for the classification.

2) Logistic Regression: is a statistical model that in its basic form uses a logistic function to model a binary dependent variable which is used due to having a categorical dichotomy as an outcome variable that violates the assumption of linearity in normal regression.it is predicted as an outcome variable that is a categorical dichotomy from one or more categorical or continuous predictor variables [23][24].

**Feature of LR**
- Forced Entry: variables should be entered simultaneously.
- Hierarchical: variables entered in the block and should be based on past research, or theory being tested.
- Stepwise: variables entered on the basis of statistical criteria and Should be used only for exploratory analysis.

3) Random Forest: RF classifier proposed by L. Breiman in 2001described as the collection of tree-structured classifiers aggregates their predictions by averaging. It has shown optimum performance in settings when the number of variables is much larger than the number of observations. In other words, it is a group of tree predictors where each tree depends on the values of a random vector sampled independently and with the same distribution for all trees in the forest. RF classifier described as the collection of tree-structured classifiers aggregates their predictions by averaging them. It has shown optimum performance in settings when the number of variables is much larger than the number of observations [25][26].
The generalization error for forests has the following features:
- Converges to a boundary when the number of trees in the forest gets large.
- Depends not only in the strength of the individual trees in the forest but also in the correlation between them.

Using a random selection of features to split each node yields error rates that compare favorably to Adaboost, as they are more robust with respect to noise. Internal estimates monitor error, strength, and correlation and these are used to show the response to increasing the number of features used in the splitting and to measure the variable significance [27][28].

Method of RF A random forest is a classifier consisting of collection of tree-structured classifiers [h(x, $\Theta_k$), k = 1, ...] where the [$\Theta_k$] are independent identically distributed random vectors and each tree casts a unit vote for the most popular class at input x [27].

4) Extra-Tree: is an ensemble machine learning algorithm that combines the predictions from many decision trees. It can often achieve a good or better performance than the random forest algorithm, although it uses a simpler algorithm to construct the decision trees used as members of the ensemble. Figure 2 illustrates examples of reply trees in tweeter nodes are tweets and edges denote a "replied to".

5) XGBoost: is an abbreviation for the extreme Gradient Boosting package. The most important factor behind the success of XGBoost is its scalability [29].It uses a tree learning algorithm and a linear model solver in order to complete ranking, classifications, and regression. It is necessary for it to be flexible to allow for users to have their own objectives [30].

**Features of XGBoost**
- Speed: it can automatically do parallel computation on Windows and Linux.
- The variety of Input data: it takes several types of input data such as Dense Matrix, Sparse Matrix, Data File, and xgb.DMatrix.
- Sparsity: it accepts sparse inputs for tree booster and linear booster, also optimized for sparse input.
- Customization and Performance: it supports customized objective function and evaluation function. Furthermore, it has better performance on several different datasets [30].

6) CatBoost: is an open-sourced gradient boosting library that handles categorical features. Categorical feature is a feature having a discrete set of values that are not necessarily comparable with each other such as user ID or name of a city.

**The features of CatBoost**
CatBoost handles categorical features and takes advantage of dealing with them during training as opposed to reprocessing time. Besides, it reduces overfitting since it uses a new schema for calculating leaf values when selecting the tree structure.
It has both CPU and GPU implementations. The GPU implementation allows for faster training. The library also has a fast CPU scoring implementation, which performs better than XGBoost and LightGBM implementations on ensembles of similar sizes [31][32].

7) Long Short-Term Memory Networks: (LSTM) networks are an extension of a recurrent neural network capable of learning order dependence in sequence prediction. The core idea is the present input taking into consideration the previous output by storing it in its memory for a short period of time. [33][34]. Correctly applying dropout to LSTM will lead to substantially reduced overfitting on a variety of tasks [35]. The network has one input layer, one hidden layer, and one output layer. The (fully) self-connected hidden layer contains memory cells and corresponding gate units.

**Features of LSTM:**
- It has been designed to tackle vanishing gradient problem.
- LSTMs handle noise, distributed representations, and continuous values such as Longtime lags.
- No need to keep a finite number of states from previous state.
- LSTMs provide a large range of parameters such as learning rates, and input and output biases so no need for adjustments.
- Reducing the complexity to update each weight to O (1) with LSTMs.

8) Convolutional Neural Network (CNN): is one of the most popular deep neural networks where its name comes from a mathematical linear operation between matrixes called convolution. It has multiple layers which have several characteristics such as convolutions layer and non-linearity layer both of them have parameters,

The CNN has optimum performance in machine learning such as the applications that deal with image data. CNN has beneficial aspects such as reducing the number of parameters in ANN, achieving abstract features when input propagates, should not have features that are spatially dependent. For instance, in face detection, the only concern is to detect faces regardless of their position in the given images and achieve abstract features when input propagates toward the deeper layers [36]

Table II illustrates examples of classification methods that used are in some researches.

### B. Generation Approaches

The goal of generation approaches is to generate CHS as the following formula [6]

$$obj = max \sum_{(c,r) \epsilon D} \log p(r|c) \qquad (1)$$

Where c is the conversation, r is the corresponding intervention response, and D is the data set. the main idea is to create a response regardless of the dialog length, language cadence, and Where c is the conversation, r is the corresponding intervention response, and D is the data set. the main idea is to create a response regardless of the dialog length, language cadence, and word imbalances. Their approach was choosing three methods Seq2Seq VAE and (RL) [6]:

- Seq2Seq: encoder consists of two bidirectional GRU layers and the decoder consists of 2 GRU layers followed by a 3-layer MLP
- Variational Auto-Encoder (VAE): has two independent linear layers followed by the encoder to calculate the mean and variance of the distribution of the latent variable separately
- Reinforcement Learning (RL): dialog generation

Although the generative models are considered an agent, it different from dialog generation. The intervention has only one tune of utterance the response will be to select a token. The state of the agent will be given by the input posts will be the state of the agent and the previously generated tokens. The result due to this difference is that the rewards with regard to ease of answering or information flow do not apply to this case, but the reward for semantic coherence does. Therefore, the reward of the agent is:

$$(rw(c,r) = \lambda 1 \log p(r|c) + \lambda 2 \log p_{back}(r|c) \qquad (2)$$

where *rw (c, r)* is the reward with regard to the conversation c and its reference response r in the dataset. *p(r|c)* denotes the probability of generating response r given the conversation c, and *pback(c/r)* denotes the backward probability of generating the conversation based on the response, which is parameterized by another generation network. The reward is a weighted combination of these two parts, which are observed after the agent finishing generating the response [6][37].

RL network is used to parameterize the probability of a response given the conversation. This network contains two Seq2Seq models. of this backbone Seq2Seq model, another Seq2Seqmodel is used to generate the backward probability it will be different in equation 2 by:

1. For *(log p(r/c)* reward will be moved by the MLE loss.
2. A learning strategy is selected for the reward of log back*(c/r)*.
3. Utilizing the baseline strategy to find the average reward where they parameterize it as a 3-layer MLP.

### V. DATA COLLECTION

In this section, we will study the data collection method in detail for each research such as the sources of collecting data, the size of the data sets, the categories of CHS, and the topics targeted to collect data. For instance, Table III illustrates the platforms that were used to collect data in each recache, table IV illustrated the topics for each research and table V illustrates the size of each data set in each research.

Qian et al. [6] collected data from both Reddit and Gab as follow:

• Reddit: to collect high-quality conversations, they referenced the list of the whiniest most low-key toxic subreddits and skipped the three subreddits that have been removed.The collection of the data was from ten subreddits: r/DankMemes, r/Imgoingtohellforthis, r/KotakuInAction, r/MensRights, r/MetaCanada, r/MGTOW, r/PussyPass, r/PussyPassDenied, r/The Donald, and r/TumblrInAction. For each of these subreddits, retrieve the top 200 hottest submissions using Reddit's API [38].

. • Gab: they collect data from all the Gab posts in October 2018 by using hate keywords to identify potentially hateful posts [38], rebuild the conversation context and clean duplicate conversations.

• The data sets consist of 5K conversations retrieved from Reddit and 12k conversations retrieved from Gab.

Mathew et al. [9] used Twitter as the platform for this study and utilized the PHEME script which allows us to collect the set of tweets replying to a specific tweet, forming a conversation. The data set contains 1290 tweet-reply pairs that considered the tweets as hate speech and their replies are CHS. The Methodology is divided into three steps: hateful tweet collection, Filtration and annotation of the hateful tweets, and Extraction and annotation of the reply tweets. The selection of hate speech tweets was decided based on at least two replies, then using the PHEME script to scrape the replies received to these hateful tweets. Observe that the 558 hate tweets received a total of 1711 replies.

Chung et al. [14] data set consists of 4,078 pairs over the three languages. They provide three types of metadata: expert demographics, hate speech sub-topic and counter-narrative type. The data set is augmented through translation (from Italian/French to English) and paraphrasing, which brought the total number of pairs to 14.988.

Tekiroglu et al. [7] use three prototypical strategies to collect HS-CN pairs have been presented recently. The Crawling (CRAWL) approach is a mix of automatic HS collection via linguistic patterns, and a manual annotation of replies to check if they are responses that counter the original hate content.

Thus, all the material collected is made of natural/real occurrences of HS-CN pairs. Crowdsourcing (CROWD) proposed that once a list of HSs is collected from SMPs and manually annotated, we can briefly instruct crowd-workers (non-expert) to write possible responses to such hate content. In this case the content is obtained in controlled settings as opposed to crawling approaches. Niche sourcing (NICHE) is idea the idea of outsourcing and collecting Counter-Narrative (CNs) in controlled settings. However, the CNs are written by NGO operators, i.e., persons specifically trained to fight online hatred via textual responses that can be considered as experts in CN production. The data collection strategies depending on the two main requirements data quantity and data quality. The key element for an effective automatic CN generation is Finding the right trade-off between the two requirements.

Garland et al. [16] used self-labeling hate and CHS groups engaged in discussions around current societal topics such as immigration and elections. Reconquista Germanica (RG)was a highly organized hate group which aimed to disrupt political discussions and promote the right-wing populist, nationalist party Alternative für Deutschland (AfD). At their peak time, RG had between 1,500 and 3,000 active members. Reconquista Internet (RI) was the counter group was formed in late April 2018 with the aim of countering RG's hateful. Active Members of RG used their public accounts to either spread hateful rhetoric, promote alt-right propaganda, or engage in directly hateful speech. Therefore, we considered the tweets sent from these accounts to be largely representative of hateful speech. RI had an estimated 62,000 registered and verified members, of which over 4,000 were active on their discord server for the first few months. However, RI quickly lost a significant number of active members, splintering into independent though cooperating smaller groups. We collected millions of tweets from members of these two groups. By building an ensemble learning system with this large corpus we were able to train highly accurate classifiers which matched human judgment. To train our classification algorithms, we collected more than 9 million relevant tweets. Of these tweets, we labeled 4,689,294 as originating from a hate account RG member tweets and 4,323,881 as originating from a CHS account RI member tweet.

For every country included in Bartlettet et al. [39] the data was collected from public Facebook Pages using an iterative process and conducted a series of analyses. This included calculating of the average interactions per Page and per post using automated API results, the format of the most popular types of data using automated API results, the type and style of the most popular types of content through manual human analysis, the types of speech occurring on different Pages using manual human analysis and the way different types of content was shared on Pages vis-à-vis users' own newsfeeds using automated analysis.

Mathew et al. [15] collected data from YouTube by collecting comments from selected video they used the YouTube comment scraper their target communities were Jews, African Americans, and LGBT. To obtain a deeper understanding they performed two steps of annotation and they achieved 95.9% agreement between the two annotators. The data set contained 4111 comments annotated as CHS and an additional 5327 comments tagged as non-CHS.

Strachan et al. [10] did not collect any data that proved the interventions to CHS which has had some success including:

1. Television programmes, such as, four episodes of a popular television series mainly focused on hate speech and incitement to violence in Kenya. An independent evaluation of the intervention concludes that the programs render citizens in areas exposed to violence more suspicious of political leaders who use inflammatory language.

2. Radio programmes, such as Dutch NGO called Radio La Benevolencija has used radio dramas, discussions, and educational programs to enable vulnerable citizens in conflict-affected countries to recognize and respond to inflammatory speech.

3. Text messages: Civil Society Organizations (CSOs) in Indonesia and Kenya used text messages to counter rumors and inflammatory speech in areas exposed to ethnic violence.

4. Monitoring hate speech, such as in Kenya, the Umati project created a database of hate speech in the run-up to the country's 2013 election.

5. Self-regulatory media systems: In Iraq, the United States Institute of Peace (USIP) supported local media stakeholders in the establishment of a self-regulatory media system, in order to reduce the prevalence of hate speech in the media.

Schieb et al. [12] set up a computational simulation model that is used to answer general questions concerning the affects that hinder or support the impact of CHS. The simulations are random, and they trade off computation time against the number of repeats done per configuration. On a normal PC, R implementation of algorithm needs about 8 hours for computing the data needed to plot one figure (220 different configurations with 50 repeats per configuration). The measured standard deviation is usually below 0.03 with this setup.

Sponholz et al. track all the articles that mention The *Rage and the Pride book*.it was directly mentioned in 74 articles in the first three months after the publication of the book. Corriere Della Sera published 43 of them and La Repubblica published 31 articles. Most of these articles were written by journalists (77%). The *Rage and the Pride* could not have turned into an international bestseller without the prior media controversy surrounding it.

Ernst et al. [13] selected eight videos of the YouTube campaign Concepts of Islam which are marked with whatIS as well as their respective user comments. Then randomly selected 155 user comments out of 5798 comments in total.

Garland et al. [17] preformed two independent data collection phases:

- Classifier Training Data Collection: They collected millions of tweets originating from approximately 3,700 known RG and RI members to train a classification system to identify hate and CHS typical of these groups, as well as neutral speech not typical of either group.
- Reply Tree Data Collection: They collected a longitudinal sample of German political conversations (or "Reply Trees") over a 4-year period during which RG and/or RI were active. In total, we collected 203,711 conversations (reply trees) that grew in response to tweets of 9,933 root accounts between January of 2013 to December of 2018. For the analysis reported here they, limited the dataset to 181,370 conversations, containing 1,222,240 tweets, which grew in response to tweets from 23 accounts for which we had continuous coverage throughout the period of January.

## VI. EVALUATION

The evaluation of CHS systems fall into two categories as mentioned in table VI. In this section, we review the most important approaches and metrics used by different research in this area.

### A. Impact of Demographics

The experiment was designed to evaluate demographic information if it has a beneficial effect on the task of counter-narrative selection/production. They selected a subsample of 230 pairs from our dataset written by 4 males and 4 females in the same age range. The subjects were presented in each pair in isolation and asked to indicate whether they would use that particular counter-narrative for the specific hate speech or not. The result was in same-gender 47% of the time where configuration declared by the operator who wrote the message and gender declared by the annotator is the same. While a different gender configuration was about 32% of the time and both the operator and annotator are different. The difference is statistically significant with $p < .001$. It concludes even though operators were following the same methods and were instructed on the same possible arguments to generate counternarratives, there is still the gender affecting produced text, and this effect contributes to the counter-narrative preference in same-gender configuration [14].

### B. Commenting Time

The main reason to use the average time essential evaluation is to decrease the time needed to produce training data for automatic CN generation by experts to have a proper pairing. Figure 3 illustrates Proportion of hate and counter speech over time. Organized counter speech (RI, blue vertical line) is followed by changes in proportions of hate and counter speech tweeted within 181,370 reply trees to 23 prominent twitter accounts, from January 2015 to December 2018. Each data point is a daily average and trends are smoothed over a one-week window.

### C. Comparison to Human Judgment

To evaluate the performance of classifiers for CHS it should have been compared with the human judge to make sure they end up with the same result. For instance, [16] they tried to collect crowdsourcing results, shown in figure 4 to double-check with an automated classifier to find out if it aligns well with human judgment.

### D. Macro-Level Effectiveness Measures and Analyzes

In Macro-Level evaluation, the impact of global events is taken into account to investigate the effectiveness of CHS. According to [17] it was demonstrated that the proportions of CHS occurred in our corpus of reply trees over time. Prior to May of 2018, when RI became functional, the relative proportions of both hate and CHS are largely in a steady state. Roughly 30 % of the discourse in our sample of political conversations during this time is hateful and only around 13 % was CHS. However, both of these time series are quite noisy, and there are several clear deviations reflecting the ebbs and flows of political discourse. Many of these deviations coincide with major events such as large-scale terrorist attacks, political rallies or speeches, and elections. After each deviation, however, the proportion of hate and CHS reverts to the earlier equilibrium, suggesting that these were so-called shock events in an otherwise steady-state system. Micro-Level analyzes the support that hate, and CHS receives through likes, subsequent replies, and retweets. Afterwards, they investigate how each hate and counter tweet steers the subsequent discussion in reply trees towards more of the same or towards opposing speech.

## VII. CHALLENGES AND OPPORTUNITIES

There are many layers of difficulty to measure the impact of CHS such as:
1. **Sequential comments**: The main obstacle to measure the CHS impact is to get access to a set of sequential comments. Processing a set of comments sequences enables the researchers to compare the comments before and after an inserted CHS in terms of sentiment analysis scores. The comments belonging to a specific post in social media might be considered as a potential resource to collect sequential comments. However, we can roughly suppose that a user reads all previous comments and submits a new comment. That's why finding a true set of sequential comments in traditional social media is rather difficult. Fortunately, enough, the rise of voice-based social media like Clubhouse can fill this gap.
2. **Definition of hate speech**: It varies universally, and no single definition is fully agreed upon in all regions. According to [41] hate speech is understood as any kind of communication in speech, writing, or behavior, that attacks or uses pejorative or discriminatory language with reference to a person or a group on the basis of who they are. In other words, based on their religion, ethnicity, nationality, race, color, descent, gender or another identity factor. However, the definition is so broad and obviously does not satisfy the criteria of being a universally accepted productive definition on several accounts [42]. In addition, there is no solid line between hate speech and the right to free speech which renders no hate speech a precise definition [43]. There is no legal definition of" hate speech" under U.S. law and hate speech can only be criminalized when it directly incites imminent criminal activity or consists of specific threats of violence targeted against a person or group [44]. Some expressions are not offensive but can be used in some way will result in a hate speech context [45][46]. As a result of a clear definition not existing while collecting data, we should look beyond the content of the speech and take into account the context, the status of the speaker, and the potential consequences of the speech act. Therefore, to evaluate the performance of classifiers for CHS we should compare it with the human judge to make sure they align with the same result.

3. **Training Data:** One of the most challenges to training data are to use a data set that is as balanced as possible because machine-learning algorithms depend highly on the quality and content of the training data [47][48]. Imbalanced data sets are when a severe skew in the class distribution leads to bias in the training data set. Although there is spread of hateful content, it is still true that this content only constitutes a small fraction of all content. The recommendation to avoid bias caused by imbalanced data according to [49] is to use four resampling methods including Random Oversampling (ROS), Synthetic Minority Technique (SMOTE), Adaptive Synthetic (ADASYN), and Random Under sampling (RUS) are used as an answer to the inequality of class distribution in a hate speech data set. With three basic machine learning classifiers i.e., Support Vector Machine, Logistic Regression, and Naive Bayes, the evaluation results show that the oversampling approach improves the accuracy and the overall performance of three classifiers. Also, biases are caused by the training public data that is concerned as biased data. For example, the toxicity scores given by the Google Jigsaw Con-versation AI, a state-of-the-art model for toxic language detection, the toxicity scores given by the Google Jigsaw Conversation AI have been discovered of giving higher toxicity scores to sentences that include female/women than male/men [50]. Moreover, any collected real-life data set contains more toxic comments regarding women, so the evaluation of toxicity becomes attached to those specific words that should only be the "neutral context [51].

## VIII. COUNTER HATE SPEECH IN THE MEDIA AND NEGATIVE IMPACT

### A. Counter Hate Speech with Negative Impact

In order to understand the negative impact of CHS, we should understand what reactance theory is. It is unpleasant motivational by external threats that arise when people experience a threat to or loss of their free behaviors [52][53]. The degree of reactance depends mainly on the importance of the threatened freedom and the perceived magnitude of the threat [54][53]. Some examples of threats to external freedom include advertisements convincing you to buy items from a supermarket that one may not need, as well as having to pay for the required tuition fees for a semester, and lastly having rules that punish the use of mobile phones in the school's district [54]. Certain people may react aggressively as they may feel that they are threatened people, with rules in place to target them. This can lead to aggressive behavior and backlash in order to speak their mind and bring attention to the issues they may face [55][56]. This is a way for these people to voice their concerns and their feelings [52][57][54]. According to reactance theory, people in social media react aggressively if they feel their freedom of speech has been touched by CHS. The negative effect not only leads to raising the tone of anger or motivates users to share hateful content but also creates groups to post more aggressive replies to the CHS which makes the case worse when it includes attacks on people's dignity, beliefs, racism

### B. Counter hate speech in the Media through Ethics

As online content continues to grow, so do hateful, racist, or homophobic comments on the internet become widely spreading. Civil society should take an action toward hate speech such as:

- CHS should take a place on civil society and social steps rather than state-initiated legal measures. The online community should work to CHS and marginalize hateful messages. Awareness of media ethics should focus on the rights and freedoms of journalists and their role in creating and promoting peaceful societies. Educating be raised on the political, social, and cultural rights of individuals and groups, including freedom of speech, and the responsibilities and social implications that come with press freedom [58][59][60].
- The awareness campaigns should emphasize the respect for the diversity of cultures and traditions. Journalists should be taught conflict-sensitive reporting skills such as avoiding making any opinion into a fact or using any expressions that lead to hate speech in any society such as terrorists, extremists, or fanatics because these words render journalists taking sides and making the negotiation with other side is impossible [61][62][63][64]. Some research emphasizes the significance of regulating social media platforms to eliminate hate speech due to being negatively impacted. Yet, given the hardness of making an excuse that the public's health, safety, and welfare are negatively impacted by hate speech on social media sites. Moreover, if this action is against freedom of speech, users should report it [65][66].

## IX. REVIEW

In this section, we summarize our in-depth investigation output.

- [9] and [15] are most frequently referred research which play central role so far in the field CHS.
- Twitter and Facebook are the most frequently investigated resource to fight hate speech.
- No voice-based social media investigated so far to practice CHS.
- Most of the previous research focus either on classification or generation. Very rare research did both tasks in addition to analysis and study.
- Most of the previous research used either expert-based or machine-based approaches. Very rare researches used
- both methods to evaluate the CHS techniques.
- Religion is the most frequently studied topic in the field of CHS.
- The XGB, LR and RF are the most frequently used methods to classify CHS instances.
- The largest collected dataset in this field is [16] with 9 million tweets used to train a classifier to discover in more than 135,000 fully resolved Twitter from 2013 to 2018.
- Only one research [17] investigated the impact of CHS in the context of time series.
- Most of the research reported positive impact for CHS with no robust supporting evidences.

## X. CONCLUSION

With the wide increase of social media platforms, hate speech has raised new concerns in society. Moreover, due to the lack of a solid definition of hate speech and the blurry line between hate speech and free freedom, hate speech is abundant on many online platforms. Hate speech cannot be banned, rather CHS plays an essential role to limit hate speech by bringing awareness to people without touching the freedom of speech principles. In this paper, we study major papers that discussed the CHS from different angles such as the method of data collection, classification methods, and the impact of CHS, etc. Also, we studied in detail both methods of CHS classification and CHS generation.

ACKNOWLEDGMENT

This research is supported by University of Houston President's Grants to Enhance Research on Racism (2020).

**Dana Alsagheer** is a PhD student in Computer Science Department at University of Houston. She got her BSc degree on 2016 and her MSc degree on 2018 both from UH CS.
Her interests include hate speech detection and counter hate speech analysis. She is currently working as Instructional Assistant in UH CS department.

**Hadi Mansourifar**, is a PhD candidate in Computer Science department, University of Houston. He received the B.S. degree in software engineering from KIAU, Iran, in 2007 and the MSc degree in software engineering from QIAU, Iran, in 2012. His PhD research focus is adversarial and non-adversarial imbalanced classification.

**Weidong Shi, is** an associate professor in the Department of Computer Science, University of Houston. He received his PhD from Georgia Institute of Technology in 2006. Before starting his academic career as faculty member at University of Houston in 2011 he worked as senior researcher in Motorola and Nokia corporations.